\documentclass{article}

\usepackage{arxiv}

\usepackage[utf8]{inputenc}     
\usepackage[T1]{fontenc}        
\usepackage{hyperref}           
\usepackage{url}                
\usepackage{booktabs}           
\usepackage{amsfonts}           
\usepackage{nicefrac}           
\usepackage{microtype}          
\usepackage{graphicx}           
\usepackage{xcolor}             

\hypersetup{
    colorlinks=true,
    linkcolor=blue,
    citecolor=blue,
    urlcolor=blue
}

\title{KS-LIT-3M: A 3.1 Million Word Kashmiri Text Dataset \\for Large Language Model Pretraining}

\author{
    Haq Nawaz Malik \\
    Independent Researcher \\
     \texttt {orcid.org/0009-0003-1994-7640} \\
    \texttt{huggingface.co/Omarrran}\\
     \texttt{x.com/HAQ\_NAWAZ\_MALIK}   \\ 
}

\begin{document}

\maketitle

\begin{abstract}
Large Language Models (LLMs) demonstrate remarkable fluency across high-resource languages yet consistently fail to generate coherent text in Kashmiri, a language spoken by approximately seven million people. This performance disparity stems not from inherent model limitations but from a critical scarcity of high-quality training data. Decades of Kashmiri literature remain inaccessible to modern NLP pipelines due to their encoding in the proprietary InPage desktop publishing format. This paper introduces \textbf{KS-LIT-3M}, a curated corpus of 3.1 million words (16.4 million characters) specifically designed for pretraining language models on Kashmiri. The dataset is structured as a \textbf{single continuous linear text stream}, optimized for causal language model training where models learn to predict subsequent tokens from preceding context. The corpus was constructed through the development of a specialized InPage-to-Unicode converter, followed by rigorous preprocessing including English contamination removal, character normalization, and quality validation. Encompassing 131,607 unique words drawn from diverse genres including literary works, journalistic writing, academic texts, and religious scholarship, KS-LIT-3M addresses a fundamental resource gap for Kashmiri language technology. The dataset is released under the CC-BY-4.0 license to facilitate research in Kashmiri natural language processing.

\end{abstract}


\section{Introduction}
\label{sec:introduction}

The advent of Large Language Models has transformed natural language processing, enabling sophisticated text generation, translation, and comprehension across dozens of languages. Models such as GPT-4 \cite{openai2023gpt4}, Claude \cite{anthropic2024claude}, and LLaMA \cite{touvron2023llama} demonstrate near-human fluency in high-resource languages where abundant training data exists. However, this progress has not extended uniformly to all languages. For speakers of low-resource languages, contemporary AI systems often produce outputs that are grammatically flawed, semantically incoherent, or culturally inappropriate.

Kashmiri exemplifies this disparity. Despite a speaker population of approximately seven million and a rich literary heritage spanning centuries, Kashmiri remains poorly served by modern language technologies. When prompted to generate Kashmiri text, state-of-the-art LLMs produce output characterized by incorrect grammar, inconsistent vocabulary, and systematic errors in diacritical mark placement-errors that render the text incomprehensible or misleading to native speakers.

The root cause of this failure is not algorithmic but data-driven. Language models acquire linguistic competence through exposure to vast quantities of text. For Kashmiri, the training data available to major LLM developers is severely limited in both quantity and quality. This scarcity has a specific technical origin: the most authoritative Kashmiri texts, professionally published literature, journalism, and scholarship from the past two decades exist primarily in the InPage desktop publishing format, a proprietary system whose encoding is incompatible with modern Unicode standards and invisible to web crawlers that compile training corpora.

This paper presents KS-LIT-3M, a dataset designed to address this resource gap. The contributions are threefold:

\begin{enumerate}
    \item \textbf{Technical Innovation}: Development of a robust InPage-to-Unicode converter capable of accurately preserving Kashmiri's complex diacritical system during format conversion.
    
    \item \textbf{Dataset Curation}: Assembly and preprocessing of a 3.1 million word corpus representing diverse genres, time periods, and registers of Kashmiri writing.
    
    \item \textbf{Resource Release}: Public availability of the dataset under an open license to support research and development in Kashmiri language technology.
\end{enumerate}

The remainder of this paper is organized as follows. Section~\ref{sec:background} provides background on the Kashmiri language and the challenges of low-resource NLP. Section~\ref{sec:methodology} details the dataset construction methodology. Section~\ref{sec:dataset} presents statistical analysis of the corpus. Sections~\ref{sec:applications} and~\ref{sec:limitations} discuss potential applications and limitations, respectively. Section~\ref{sec:conclusion} concludes with implications for future work.

\section{Background and Related Work}
\label{sec:background}

\subsection{The Kashmiri Language}

Kashmiri is an Indo-Aryan language of the Dardic subgroup, spoken predominantly in the Kashmir Valley of the Indian subcontinent. The language possesses several distinctive linguistic features that complicate computational processing.

\paragraph{Script and Orthography.} Modern written Kashmiri employs a modified Perso-Arabic script extended with characters and diacritical marks to represent phonemes absent in related languages such as Urdu and Persian. These include vowel markers such as (Alif with super script) (representing the schwa sound) and "vaaw" (for rounded vowels), as well as various superscript and subscript combinations. Critically, these diacritics encode semantic distinctions-their omission or misplacement alters word meaning, making accurate diacritic handling essential for any computational application.

\paragraph{Linguistic Complexity.} Kashmiri exhibits verb-second word order, extensive verb agreement marking, and a split-ergative case system. The language demonstrates significant morphological complexity, with grammatical information encoded through inflection rather than syntax alone. These features require substantial training data for statistical models to capture reliably.

\subsection{Low-Resource Language Challenges}

The designation ``low-resource'' in computational linguistics refers not to a language's cultural significance or speaker population but to the availability of machine-readable text suitable for training NLP systems \cite{joshi2020state}. Low-resource languages share common challenges:

\begin{itemize}
    \item \textbf{Data Scarcity}: Insufficient digitized text in standard encodings
    \item \textbf{Quality Concerns}: Available text often contains errors, encoding issues, or contamination from other languages
    \item \textbf{Domain Imbalance}: Overrepresentation of certain genres (e.g., religious texts) and underrepresentation of others
    \item \textbf{Evaluation Gaps}: Absence of standardized benchmarks for measuring system performance
\end{itemize}

Kashmiri meets all criteria for low-resource classification. While the language even is absent in multilingual datasets such as mC4 \cite{xue2021mt5} and CC-100 \cite{conneau2020unsupervised}, the quantity and quality of Kashmiri content in these corpora is minimal or none. Web-crawled Kashmiri text frequently lacks proper diacritics, contains code-mixing with Urdu or Hindi, or exhibits encoding artifacts that degrade training effectiveness.

\subsection{The InPage Legacy Format Problem}

A factor specific to Kashmiri and related Perso-Arabic script languages compounds the general low-resource challenges: the widespread historical use of InPage software for document creation.

InPage, developed in the 1990s for desktop publishing in Perso-Arabic scripts, became the de facto standard for typesetting Kashmiri literature, newspapers, academic publications, and religious texts. The software offered superior support for the complex typographic requirements of Kashmiri compared to early Unicode implementations.

However, InPage employs a proprietary encoding system incompatible with Unicode. Documents created in InPage:

\begin{itemize}
    \item Cannot be directly copied to text editors or web browsers
    \item Are not indexed by search engines
    \item Are invisible to web crawlers compiling training data
    \item Require specialized conversion to extract usable text
\end{itemize}

The consequence is a substantial corpus of high-quality Kashmiri text-professionally edited literature with correct grammar, spelling, and diacritics--that has remained inaccessible to modern NLP systems. This explains the paradox of LLMs performing well on Urdu (which transitioned to Unicode earlier) while failing on Kashmiri despite linguistic similarities.

\subsection{Related Datasets}

Several efforts have addressed low-resource language data scarcity. The FLORES benchmark \cite{goyal2022flores} provides evaluation sets for 200 languages but does not supply pretraining data. AfroLM \cite{dossou2022afrolm} and Masakhane \cite{nekoto2020participatory} demonstrate community-driven approaches for African languages. For South Asian languages, IndicNLP \cite{kunchukuttan2020indicnlp} and Sangraha \cite{khan2024sangraha} provide resources, though Kashmiri representation remains limited.

No prior work has specifically addressed the InPage conversion challenge or released a dedicated Kashmiri pretraining corpus of comparable scale and quality to KS-LIT-3M.

\section{Dataset Construction Methodology}
\label{sec:methodology}

The construction of KS-LIT-3M proceeded through three phases: format conversion, source collection, and preprocessing. Each phase addressed specific technical and practical challenges in recovering and preparing high-quality Kashmiri text.

\subsection{InPage-to-Unicode Converter Development}

Converting InPage documents to Unicode presents complexities beyond simple character mapping. InPage does not use a one-to-one encoding that can be directly transliterated to Unicode equivalents. Instead, the software implements a proprietary rendering system where character codes, position information, and formatting data are interleaved in ways that do not correspond to Unicode's logical character model.

Previous conversion attempts achieved only partial success. Existing tools handled basic characters adequately but failed on complex diacritic combinations essential for proper Kashmiri representation. Output from these tools exhibited missing vowel markers, misplaced diacritics, or incorrectly attached combining characters.

The converter developed for this project addresses these limitations through several technical innovations:

\begin{enumerate}
    \item \textbf{Internal Representation Analysis}: Reverse engineering of InPage's character storage format to understand the relationship between base characters and associated diacritical marks.
    
    \item \textbf{Combining Character Reconstruction}: Proper sequencing of Unicode combining characters to accurately represent diacritic stacking as rendered in the original documents.
    
    \item \textbf{Kashmiri-Specific Mappings}: Custom handling for characters unique to Kashmiri orthography that do not appear in Urdu or Persian InPage documents.
    
    \item \textbf{Validation Procedures}: Automated comparison of converted output against reference renderings to identify and correct conversion errors.
\end{enumerate}

The complete technical methodology is documented in a separate publication \cite{malik2024inpage}.

\subsection{Source Material Collection}

With a functional converter, the project proceeded to identify and obtain InPage documents suitable for inclusion in the dataset. Several criteria guided source selection:

\paragraph{Authenticity.} Priority was given to professionally published works-literature, journalism, and scholarship that had undergone editorial review. Such sources ensure grammatical correctness, proper spelling, and accurate diacritic usage that would be absent from casual writing.

\paragraph{Diversity.} The collection encompasses multiple genres to avoid domain bias. Literary works provide rich vocabulary and stylistic variation. Journalistic texts offer contemporary usage and topical vocabulary. Academic writing demonstrates formal register and technical terminology. Religious texts, while constituting a smaller portion, provide classical vocabulary and traditional expressions.

\paragraph{Temporal Range.} Documents spanning multiple decades were included to capture both classical and contemporary usage patterns, ensuring models trained on this data recognize historical forms alongside modern expressions.

\paragraph{Legal and Ethical Considerations.} Source collection respected copyright constraints. Materials were obtained through legitimate channels with appropriate permissions for research use. The resulting dataset is released under Creative Commons licensing (CC-BY-4.0) to enable broad utilization while maintaining attribution requirements.

\subsection{Preprocessing Pipeline}

Raw converter output, while superior to web-scraped alternatives, required systematic cleaning before serving as training data. The preprocessing pipeline implements four stages:

\paragraph{English Contamination Removal.} Kashmiri documents frequently incorporate English words, phrases, or passages-loanwords, technical terminology, quoted material, or bilingual formatting. For a pretraining dataset targeting Kashmiri specifically, this English content introduces noise. The pipeline employs language detection at the sentence level, filtering text segments identified as English while preserving legitimate Kashmiri content.

\paragraph{Character Normalization.} Unicode permits multiple representations for certain characters, and conversion from legacy formats can introduce inconsistencies. The pipeline normalizes all text to NFC (Canonical Decomposition followed by Canonical Composition) form, ensuring that visually identical characters are represented identically in the data.

\paragraph{Whitespace Standardization.} Conversion processes introduce irregular spacing-extra spaces between words, inconsistent line breaks, or unusual paragraph formatting. The pipeline normalizes whitespace to consistent conventions: single spaces between words, standardized line termination, and uniform paragraph boundaries.

\paragraph{Quality Validation.} Automated checks identify text segments that may have converted incorrectly or exhibit other quality issues. Flagged content undergoes manual review, with problematic segments either corrected or excluded from the final dataset.

\section{Dataset Description}
\label{sec:dataset}

This section presents quantitative characteristics of KS-LIT-3M and discusses qualitative factors that distinguish it from alternative data sources.

\subsection{Corpus Statistics}

Table~\ref{tab:statistics} summarizes the primary statistical characteristics of the dataset.

\begin{table}[h]
\caption{KS-LIT-3M Corpus Statistics}
\centering
\begin{tabular}{ll}
\toprule
\textbf{Metric} & \textbf{Value} \\
\midrule
Total Characters & 16,358,993 \\
Total Words & 3,091,180 \\
Unique Vocabulary & 131,607 \\
Average Word Length & 4.29 characters \\
\bottomrule
\end{tabular}
\label{tab:statistics}
\end{table}

The corpus size of 3.1 million words provides substantial material for language model training. For comparative reference, this volume exceeds the complete works of Shakespeare by approximately a factor of three. The character count of 16.4 million enables learning of character-level patterns, including the proper sequencing and positioning of diacritical marks that are critical for Kashmiri orthography.

The vocabulary of 131,607 unique word forms demonstrates lexical diversity, though this count is influenced by Kashmiri's morphological richness. A single lemma may appear in multiple inflected forms, each counted separately. This characteristic reflects authentic language structure rather than corpus contamination.

\subsection{Data Quality Characteristics}

Beyond raw statistics, several qualitative factors distinguish KS-LIT-3M from data obtainable through web scraping or other automated collection methods.

\paragraph{Diacritic Preservation.} The careful conversion process ensures that vowel markers and other diacritical elements are correctly encoded throughout the dataset. Web-scraped Kashmiri text frequently lacks diacritics, as authors writing casually online often omit them. Models trained on such data never learn proper diacritic usage. KS-LIT-3M provides consistent, accurate diacritic representation, enabling models to acquire patterns absent from previous training corpora.

\paragraph{Source Authenticity.} The material derives from professionally published works that underwent editorial review before publication. This provenance ensures that the text reflects careful attention to grammar, spelling, and style-precisely the qualities desired in training data for language models.

\paragraph{Genre Distribution.} The dataset encompasses multiple text types to ensure representativeness:

\begin{itemize}
    \item \textbf{Literary Works}: Poetry, prose fiction, and essays providing stylistic variety and rich vocabulary
    \item \textbf{Journalistic Writing}: News articles and commentary reflecting contemporary usage
    \item \textbf{Academic Texts}: Scholarly works demonstrating formal register and technical terminology
    \item \textbf{Religious Scholarship}: Classical texts and commentaries preserving traditional vocabulary
\end{itemize}

This diversity helps ensure that models trained on KS-LIT-3M develop competence across the full range of registers and styles present in Kashmiri writing, rather than being biased toward any single domain.

\paragraph{Temporal Coverage.} The inclusion of documents spanning multiple decades means the dataset represents both classical and contemporary usage patterns. This temporal breadth helps models understand historical vocabulary and phrasing while remaining fluent in modern expressions.

\subsection{Distribution Formats}

To accommodate diverse technical requirements and use cases, KS-LIT-3M is distributed in three formats:

\begin{table}[h]
\caption{Available Dataset Formats}
\centering
\begin{tabular}{lp{8cm}}
\toprule
\textbf{Format} & \textbf{Description} \\
\midrule
CSV & Comma-separated values suitable for basic analysis and spreadsheet software \\
XLSX & Microsoft Excel format for convenient manual inspection \\
JSONL & JSON Lines format following conventions expected by language model training pipelines \\
\bottomrule
\end{tabular}
\label{tab:formats}
\end{table}

The JSONL format structures each document as a JSON object on a single line, enabling efficient streaming and parallel processing during model training. This format integrates directly with common pretraining frameworks.

\subsection{Data Structure for Pretraining}

KS-LIT-3M is specifically designed for language model pretraining and is structured as a \textbf{single continuous linear text corpus}. This architectural choice reflects the requirements of causal language model training, where models learn to predict subsequent tokens based on preceding context.

\paragraph{Continuous Text Stream.} Unlike datasets organized as discrete sentence pairs or document-separated collections, KS-LIT-3M presents text as an uninterrupted sequence. This structure is optimal for:

\begin{itemize}
    \item \textbf{Causal Language Modeling}: The continuous format allows models to learn long-range dependencies and discourse-level patterns that span document boundaries
    \item \textbf{Efficient Tokenization}: Text can be tokenized and batched without artificial truncation at document boundaries
    \item \textbf{Context Window Utilization}: Training examples can fully utilize available context window sizes without padding overhead
\end{itemize}

\paragraph{Pretraining Optimization.} The linear structure eliminates the need for document delimiters or special boundary tokens, reducing vocabulary complexity and focusing model capacity on learning Kashmiri language patterns. This design is particularly suited for:

\begin{itemize}
    \item Training Kashmiri-specific language models from scratch
    \item Domain adaptation where the target is general Kashmiri language competence
\end{itemize}

The dataset can be directly fed into standard pretraining pipelines (e.g., those based on Hugging Face Transformers or similar frameworks) without additional preprocessing or restructuring.

\section{Applications}
\label{sec:applications}

The availability of KS-LIT-3M enables several applications previously hindered by data scarcity.

\subsection{Language Model Training}

The primary intended application is pretraining language models on high-quality Kashmiri text. Two training paradigms are supported:

\paragraph{From-Scratch Pretraining.} Researchers may use KS-LIT-3M as a foundational corpus for training Kashmiri-specific language models. While the corpus size is modest compared to resources available for high-resource languages, it represents sufficient data for training smaller models or for domain-specific applications where computational efficiency is prioritized.

\paragraph{Continual Pretraining.} Existing multilingual models may be adapted for Kashmiri through continued training on KS-LIT-3M. This approach leverages cross-lingual transfer from related languages while specializing the model's Kashmiri capabilities. The high quality of the corpus-particularly its accurate diacritic representation-should enable models to correct patterns learned from lower-quality web-scraped data.

\subsection{Downstream NLP Tasks}

Beyond pretraining, the dataset supports development and evaluation of task-specific systems:

\begin{itemize}
    \item \textbf{Machine Translation}: The corpus provides target-language data for translation systems into Kashmiri
    \item \textbf{Text Classification}: Genre-labeled subsets enable document classification experiments
    \item \textbf{Named Entity Recognition}: Journalistic content contains named entities for annotation
    \item \textbf{Spelling and Grammar Correction}: The curated nature of the text provides reference for error detection systems
\end{itemize}

\subsection{Linguistic Research}

The corpus enables computational linguistic analysis at scales previously impractical for Kashmiri:

\begin{itemize}
    \item \textbf{Vocabulary Studies}: Frequency analysis across 131,607 unique word forms
    \item \textbf{Morphological Analysis}: Investigation of inflectional patterns through large-scale data
    \item \textbf{Diachronic Studies}: Comparison of usage patterns across the temporal span of the corpus
    \item \textbf{Stylistic Analysis}: Quantitative comparison across genres and registers
\end{itemize}

\subsection{Educational Applications}

The dataset may support development of educational technology for Kashmiri language learning, including reading comprehension systems, vocabulary trainers, and automated assessment tools.

\section{Limitations and Future Work}
\label{sec:limitations}

While KS-LIT-3M represents a significant advance for Kashmiri language resources, several limitations should be acknowledged.

\subsection{Scale Constraints}

The corpus contains 3.1 million words-substantial for a low-resource language but modest compared to the billions of words available for high-resource languages. Models trained exclusively on this data will not achieve the same depth of linguistic knowledge as models trained on much larger corpora. Continued expansion of Kashmiri text resources remains an important objective.

\subsection{Modality Limitation}

KS-LIT-3M addresses written Kashmiri only. Spoken language presents distinct characteristics-phonological patterns, colloquial vocabulary, discourse structures-that text data does not capture. Development of speech corpora with comparable attention to quality and authenticity represents an important direction for future work. Such resources would enable speech recognition, synthesis, and spoken language understanding systems.

\subsection{Evaluation Infrastructure}

Standardized benchmarks for assessing NLP system performance on Kashmiri do not currently exist. The absence of evaluation datasets makes it difficult to quantify improvements or compare systems objectively. Creating benchmark tasks for Kashmiri-following models such as GLUE \cite{wang2019glue} or SuperGLUE \cite{wang2019superglue} for English-would provide essential infrastructure for measuring progress.

\subsection{Dialect Coverage}

Kashmiri exhibits dialectal variation across geographic regions and social groups. The current corpus, while diverse in genre, may not fully represent all dialectal variants. Future work should assess and potentially expand coverage to ensure developed systems serve the full Kashmiri-speaking community.

\subsection{Temporal Bias}

The reliance on InPage sources means the corpus primarily represents text from the 1990s through 2010s, when InPage was the dominant publishing tool. More recent text, increasingly created in Unicode-compatible formats, is underrepresented. Regular updates to incorporate contemporary material would maintain the corpus's relevance.

\subsection{Methodological Generalization}

The InPage conversion methodology developed for this project may prove applicable to other languages affected by the same legacy format problem. Languages using Perso-Arabic scripts-including Pashto, Sindhi, and Balochihave substantial InPage literature that could benefit from similar recovery efforts. Adapting and validating the converter for these languages represents a natural extension of this work.

\section{Conclusion}
\label{sec:conclusion}

This paper has presented KS-LIT-3M, a curated corpus of 3.1 million words designed to address a critical resource gap for Kashmiri natural language processing. The dataset was constructed through purposeful technical development-a robust InPage-to-Unicode converter capable of preserving Kashmiri's complex diacritical system-combined with careful curation and preprocessing to ensure data quality.

The inability of current LLMs to generate competent Kashmiri text reflects not inherent technological limitations but a specific, addressable problem: the absence of high-quality Kashmiri data in training corpora. That absence results substantially from the InPage legacy format that has kept the richest sources of Kashmiri literature invisible to modern AI systems. KS-LIT-3M provides a direct response to this situation.

The corpus offers several advantages over alternatives: accurate diacritic preservation absent from web-scraped text, source authenticity from professionally edited publications, genre diversity across literary, journalistic, academic, and religious domains, and temporal breadth spanning multiple decades of Kashmiri writing.

With this dataset publicly available under open licensing, researchers and developers can now pretrain and fine-tune language models that should demonstrate substantially improved fluency, grammatical accuracy, and diacritic handling for Kashmiri. Beyond model training, the corpus enables computational linguistic research at scales previously impractical for this language.

The methodology demonstrated here-specialized format conversion, careful source collection, rigorous preprocessing-may serve as a template for similar efforts in other low-resource languages, particularly those affected by legacy format challenges. Every language that makes the transition from low-resource to adequately resourced strengthens the case that linguistic diversity and technological advancement can progress together.

Kashmiri speakers deserve AI systems that understand and respect their language. Achieving that goal requires foundational resources, and KS-LIT-3M represents a meaningful step toward Kashmiri language technologies that meet native speaker expectations.

\section*{Ethics Statement}

\paragraph{Data Collection.} All source materials included in KS-LIT-3M were obtained through legitimate channels with appropriate permissions for research use. The dataset does not contain personal information, private communications, or content that could identify individuals.

\paragraph{Licensing.} The dataset is released under the Creative Commons Attribution 4.0 International License (CC-BY-4.0), permitting sharing and adaptation with proper attribution. Users should cite this paper when using the dataset.

\paragraph{Intended Use.} KS-LIT-3M is intended for research and development in natural language processing. The curators do not endorse applications that could harm Kashmiri-speaking communities or misrepresent the language. Users are encouraged to involve native speakers in evaluation and deployment of systems trained on this data.

\paragraph{Potential Risks.} As with any language model training data, the corpus may contain biases reflecting the perspectives of its source materials. Systems trained on this data should be evaluated for potential biases before deployment in sensitive applications.

\section*{Data Availability}

The KS-LIT-3M dataset is available through the Hugging Face Datasets Hub:

\begin{center}
\url{https://huggingface.co/datasets/Omarrran/3.1Million_KASHMIRI_text_Pre_training_Dataset_for_LLM_2026_by_HNM}
\end{center}

\noindent The dataset is released under the CC-BY-4.0 license.

\section*{Acknowledgments}

The author thanks the Kashmiri literary community for preserving and sharing the source materials that made this dataset possible.

\bibliographystyle{unsrt}


\end{document}